# Task presentation and execution paths shape LLM mathematical performance

Min Huang and Mingyan Liu

minhuang1@link.cuhk.edu.cn and msfocus@gmail.com

## Abstract

Mathematical performance depends on how a language model receives information, uses generated text and submits an answer. We study these dependencies through paired comparisons that separate submission requirements, draft reuse and candidate availability. On 65 development problems, an instruction permitting continued derivation produced no valid final answers in 390 short-output calls; a single-line submission instruction produced 390 format-valid outputs, including 389 parseable answers. Under the fixed submission protocol, 198 problems with failed initial submissions were answered correctly in 9.09% of cases with the question alone, 33.84% with its draft, 46.46% with candidates, and 59.60% with both. Drafts added 13.13 percentage points with candidates fixed, and candidates added 25.76 points with drafts fixed; both effects were significant after Holm correction and were subadditive on the accuracy scale. Retaining a trailing excerpt outperformed retaining a leading excerpt of similar average length by 10.10 points (Holm-adjusted $p = 0.027$). A second completion model showed the same directions of input effects on the fixed materials. In a cohort with parseable initial submissions, candidates added no correct answers when the previous draft was supplied, but added 27 points when it was absent. These results establish a conditional value of generated text: a failed submission can leave useful material, whose contribution depends on what else is supplied and what is retained. A single answer's success or failure therefore does not fully characterize the value produced during the answering process.

## 1. Introduction

A language model can produce a substantial mathematical derivation without delivering a usable final answer. The attempt is scored as unsuccessful, yet its text may contain calculations or candidate solutions that help a later request. A user must then decide what to carry forward and what to ask for next. Treating the attempt only as a failure gives little guidance for those decisions: it does not reveal whether the generated work remains useful, whether candidates supply the missing assistance, or whether a shorter record preserves that assistance.

This question connects task presentation to the model's execution path. Candidate answers change the available information; output instructions determine whether a call develops a derivation or submits a result; and the material passed between calls determines what the next request can use. Accuracy measures the outcome of this entire process. To explain a change in performance, the contributions of these choices must be separated.

We organize the analysis around input presentation, text generation, intermediate-material reuse and answer submission. Three experiments build the argument. First, a development comparison establishes a submission instruction that produces usable final outputs within a short allowance. Second, a factorial experiment on the same 198 problems independently varies draft and candidate availability, distinguishing the value of unsuccessful work from assistance supplied by the candidates.

Third, a retention comparison asks whether that value survives when only the beginning or end of the draft is passed forward. Supplementary comparisons examine a second completion model and changes in the initial-submission status, recovery allowance and instruction language.

The central finding is that generated text has a conditional downstream value. With candidates fixed, preserving an unsuccessful draft improves accuracy by 13.13 percentage points. With drafts fixed, candidates add 25.76 points, and the two contributions are subadditive. Retaining the end of a draft produces 20 more correct answers than retaining its beginning at a similar average length. These comparisons link performance to specific decisions in the answering process: what information is supplied, which generated material survives, and what the next call is asked to deliver.

The contribution is a controlled characterization of the downstream value of failed-submission text. Prior work has studied answer extraction from partial traces and the reuse of intermediate results (Wang and Zhu 2026; Ballon et al. 2026; Liu et al. 2026). We connect these questions through paired interventions that separate draft availability from candidate availability, then compare which part of the draft is retained. This approach links each measured difference to a choice a user can make in a subsequent call. It also offers a way to evaluate answering workflows: assess what each input contributes under a common submission protocol, rather than attributing the final score to the model or to the retained draft alone.

## 2. Related work

Intermediate text can be studied at several points in a model's answering process. Chain-of-thought prompting asks whether eliciting intermediate steps improves reasoning. Faithfulness studies ask whether subsequent answers depend on the stated reasoning, including through interventions that alter it. These questions establish why the content of a derivation matters, but they also distinguish the usefulness of text from an explanation of the model's internal computation (Wei et al. 2022; Lanham et al. 2023).

Partial-trace recovery is especially close to our setting. Wang and Zhu (2026) compare free continuation and forced extraction from reasoning prefixes, including comparisons with no prefix. Ballon et al. (2026) truncate generated traces at token percentiles and reintroduce them to the same or a different model, probing answer-choice distributions on GPQA Diamond and MMLU-Pro. These studies establish the importance of when and how generated text is read. We focus on drafts left by attempts that reached their output limit without a parseable final submission. Under a common mathematical-value output requirement, we independently vary draft and candidate availability, then compare leading and trailing retention at similar average lengths.

Options require their own control because they change the information available to the model. Raman et al. study how and when options are exposed and how models exploit their contents. In a recovery setting, attributing the success of a draft-plus-options request entirely to the draft would therefore conflate two inputs. Our factorial comparison separates them while requiring the same mathematical-value output in every condition (Raman et al. 2026).

Multi-call methods differ in what they carry forward. Self-Refine uses feedback to revise an output, and Reflexion carries linguistic feedback into later attempts (Madaan et al. 2023; Shinn et al. 2023). Liu et al. (2026) study the reliability and computational cost of intermediate-result reuse in composite mathematics and arithmetic, including how retained answers and reasoning affect subsequent subproblems. Our unit of analysis is another submission for the same problem using its original unfinished text. The factorial controls measure what that text adds alongside candidates; the retention controls measure how its value changes when only its beginning or end is supplied.

# 3. Method

## 3.1 Observable stages of mathematical answering

We distinguish four operations in the interaction with the model. **Input presentation** supplies the question and, when available, candidate answers. **Text generation** produces a working record under an output allowance. **Material reuse** passes all, part or none of that record to a fresh call. **Answer submission** specifies the final mathematical value to be returned.

This separation makes the comparisons interpretable. Candidate availability changes while the requested output remains a mathematical value, so it does not switch the task to producing an option label. Drafts are retained exactly as generated and may contain partial calculations, answer cues or errors. Their value is measured by the change in subsequent accuracy when they are supplied or removed. Format validity, parseability and mathematical correctness are recorded separately.

## 3.2 An answering procedure with material reuse

Algorithm 1 makes the transition from unfinished work to a final submission explicit. It checks whether the initial output contains a parseable answer and, if not, supplies the existing text to a fresh request for that answer. This trigger can be applied without knowing the reference solution. Reference-answer correctness is used only to evaluate the resulting submissions. The experiments examine the recovery branch by varying the materials passed through it.

**Algorithm 1. An answering path with draft reuse and explicit submission**

```
Input: problem q, initial output d, available options C,
    model F, final-answer parser P, recovery output limit B

1. If P(d) contains a valid final submission, return it.
2. Otherwise, create a fresh request containing:
    the original problem q;
    the full generated text d as working material;
    the available options C, or an empty options field;
    an instruction to submit only the final mathematical value.
3. Generate one response with model F and output limit B.
4. Return its parsed answer, or mark the submission unresolved.
```

The algorithm specifies the workflow whose recovery branch we evaluate. Its trigger is an absent parseable submission, so it does not attempt to correct valid but incorrect initial answers. Candidate sets are constructed from reference answers and distractors. A full draft is all text produced before the initial output limit, not a completed solution.

## 3.3 Comparison design and attribution

The primary design crosses draft visibility with option visibility. The four requests contain the question alone, the question and draft, the question and options, or all three. Each problem receives every request, so differences in problem difficulty cannot explain the comparison between conditions. The question-only condition measures what another call can achieve under the same brief-submission protocol. The options-without-draft condition measures recovery from candidate information alone. Together, these controls distinguish the contribution of retained text from the opportunity to answer again and from assistance supplied by options.

Comparing the two conditions with options holds candidate information fixed and estimates the draft's incremental contribution. Comparing the two conditions with a draft holds prior text fixed and estimates the contribution of options. The interaction asks whether either gain changes when the other input is present. All four conditions require a mathematical value, keeping the deliverable constant as the available information changes. Fresh sessions ensure that a removed input is not still available through conversation history.

The subsequent retention comparison asks which version of the draft supports submission. It keeps the question, candidates and final-answer instruction fixed while supplying the full draft, a leading excerpt, a trailing excerpt or no draft. Full versus empty measures the benefit of reuse; full versus leading measures the cost of retaining only the beginning. Leading versus trailing compares two retention choices with similar average token budgets, asking whether the quantity of retained text is enough to account for their performance. The comparison changes the selected content along with its original location in the draft, so it evaluates a retention policy rather than isolating position as an internal model mechanism.

# 4. Experimental setup and results

## 4.1 Model, samples and evaluation

The primary experiments use a fixed Qwen3.5-4B checkpoint and tokenizer, greedy decoding and disabled hidden thinking. Recovery calls use fresh sessions, with earlier text supplied explicitly. They request one final mathematical value within 32 tokens; unused input fields remain empty. Supplementary comparisons change the completion model, allowance or instruction language as specified in Section 4.5 and Appendix F.

The primary cohort comes from an existing 800-problem collection spanning seven mathematical subjects. Among 201 initial attempts without a strictly parseable final answer, an automated audit identifies three explicit conclusions in unsupported notation and classifies the remaining 198 outputs, which reached the 1,024-token limit, as intermediate-only. The 198 form the primary analysis; all 201 are included in a sensitivity analysis. This automated classification concerns submission status and does not exclude answer cues within the drafts.

The primary cohort and the 65 development problems are disjoint by identifier and normalized question hash. The source collection had previously been used in the project. The two primary contrasts were specified in a local study plan before recovery calls; the retention and supplementary analyses were developed subsequently. The reported primary accuracies are recovery rates conditional on initial submission failure, rather than accuracy estimates for the full 800-problem collection. Appendix A documents the archived sample selection and runtime provenance.

Correctness is scored by mathematical equivalence to the reference answer. All assigned problems remain in the denominator, with missing, truncated or unparseable submissions scored as unsuccessful. The two primary contrasts use two-sided exact McNemar tests with Holm correction and 95% intervals from 10,000 item-level paired bootstrap resamples. The accuracy-scale interaction is a secondary one-sample t-test on item-level contrasts. Retention uses a separate six-comparison Holm family; each supplementary factorial analysis uses a two-contrast family, with interaction p-values reported unadjusted. Appendices A, B and F provide execution and scoring details.

## 4.2 Making final submission possible

On 65 development problems, six conditions with a 32-token output allowance produced 390 calls under an instruction permitting continued derivation. None delivered a valid final answer. After the instruction was changed to require a single-line final answer without explanation, all 390 calls produced format-valid outputs and 389 were mathematically parseable. The original requests used the short allowance to begin or continue a derivation before reaching submission.

This comparison concerns successful delivery, not mathematical correctness. It shows that the output requirement and available allowance jointly determine whether a final answer is observed. We fixed the revised submission instruction for the subsequent experiments so that differences between draft and option conditions could be interpreted under a common output requirement. Appendix A gives the executed prompt.

## 4.3 Measuring what drafts and candidates contribute

The most direct alternative explanation is that a second call can succeed from the question and options alone. The no-draft condition measures that possibility. Retaining the draft nevertheless produces an additional improvement (Table 1).

**Table 1. Accuracy on 198 problems with unsuccessful initial submissions.**

| Recovery input | Correct / 198 | Accuracy |
|---|---:|---:|
| Question only | 18 | 9.09 per cent |
| Question + draft | 67 | 33.84 per cent |
| Question + options | 92 | 46.46 per cent |
| Question + draft + options | 118 | 59.60 per cent |

With options held fixed, the draft changes 39 problems from incorrect to correct and 13 from correct to incorrect. The net gain is 26 correct answers, or 13.13 percentage points. Thus, the draft improves submission beyond what the question and candidates alone provide (Table 2).

**Table 2. Primary paired effects.**

| Change to the recovery input | Gain (percentage points) | 95 per cent interval | Holm-adjusted p |
|---|---:|---|---:|
| Retain draft, options fixed | 13.13 | [6.06, 19.70] | 0.000410 |
| Add options, draft fixed | 25.76 | [18.69, 32.83] | $1.13 \times 10^{-11}$ |

Failure to submit an answer therefore does not exhaust the value of the generated text. The gain is an average over both helpful and harmful drafts: the 13 reversals show that retention can also steer individual responses away from a correct answer. Including the three identified explicit-but-unparseable conclusions yields a draft gain of 13.43 percentage points and an option gain of 26.37 percentage points (95 per cent interval [19.40, 33.33], Holm-adjusted $p = 3.3 \times 10^{-12}$), so those cases do not account for either result.

**Contribution of options and interaction.**

With the draft held fixed, options increase accuracy from 33.84 per cent to 59.60 per cent. They correct 56 problems and worsen five. Since both conditions output a mathematical value, this gain arises from changing the information presented rather than switching the deliverable to an option label.

The four inputs also reveal how their benefits combine. Options add 37.37 percentage points without a draft and 25.76 points with one. Conversely, the draft adds 24.75 points without options and 13.13 with them. The resulting accuracy-scale interaction is −11.62 points, with a 95 per cent interval of [−19.19, −4.04] and secondary-test p = 0.00301.

Each input provides a smaller marginal gain when the other is present. The negative interaction therefore identifies a dependency between the inputs on the accuracy scale: their observed benefits cannot be added as independent improvements. This is an average relationship, with both helpful and harmful combinations at the item level (Appendix E). Together, the paired contrasts establish that failed submissions leave useful text and that its contribution depends on the other information supplied.

## 4.4 Preserving useful material when shortening a draft

The draft effect establishes that prior text can help. It does not establish that any excerpt will preserve that help. We therefore compare full drafts with their beginnings and ends, using the same question, candidates and submission instruction in every condition.

This follow-up analysis covers all 198 primary problems. It combines 240 existing records from four retention conditions on 60 problems with 552 new calls on the remaining 138 problems. The leading and trailing excerpts contain an average of 583.8 and 581.7 tokens, respectively, compared with 1,023.9 tokens for the full draft. Boundaries follow nearby sentences or paragraphs where possible. Thus, the excerpts have similar average lengths, although their lengths are not identical within every problem. Appendix C describes construction and record provenance.

**Table 3. Recovery accuracy under four material-retention policies.**

| Retained material | Correct / 198 | Accuracy |
|---|---:|---:|
| Full generated draft | 118 | 59.60% |
| Trailing excerpt | 116 | 58.59% |
| Leading excerpt | 96 | 48.48% |
| No draft | 92 | 46.46% |

The full-draft and no-draft conditions have the same item-level correctness outcomes as their counterparts in the factorial experiment. Relative to retaining the full draft, retaining its beginning loses 22 correct answers on balance. The trailing excerpt answers 20 more problems correctly than the leading excerpt, despite their similar average lengths (Table 4).

**Table 4. Paired effects of retention policy.**

| Comparison | Difference in percentage points | 95% interval | Holm-adjusted p |
|---|---:|---|---:|
| Full − leading | 11.11 | [4.55, 17.68] | 0.00821 |
| Full − trailing | 1.01 | [−2.53, 5.05] | 1.00000 |
| Trailing − leading | 10.10 | [3.03, 17.17] | 0.02722 |
| Trailing − no draft | 12.12 | [4.55, 19.70] | 0.00893 |
| Leading − no draft | 2.02 | [−3.54, 8.08] | 1.00000 |
| Full − no draft | 13.13 | [6.06, 20.20] | 0.00246 |

Intervals use item-level paired percentile bootstrapping; exact McNemar p-values are Holm-adjusted across six comparisons. Separate resampling explains the small interval-endpoint differences from the primary analysis for full versus no draft.

Trailing rather than leading retention changes 37 answers from incorrect to correct and 17 from correct to incorrect. Against no draft, the corresponding counts are 41 and 17. Thus, the trailing policy improves on both a similar-length excerpt and answering without prior material. The full draft answers two more problems correctly than the trailing excerpt; their difference interval [−2.53, 5.05] does not establish equivalence.

The consequence for reuse is specific: preserving a roughly equal quantity of text does not preserve equal downstream value. In this setting, retaining the end is more effective than retaining the beginning. These policies change both the content and its source location; the experiment identifies their performance difference. Exploratory attempts to attribute that difference to particular content are reported in Appendix E.

## 4.5 The contribution of information across answering conditions

The main experiment concerns one completion model and drafts left by failed initial submissions. We next examine whether the input effects extend to a second completion model and how they change when a parseable initial answer is already available. Table 5 separates these two questions from checks on the output allowance and prompt language.

**Table 5. Supplementary accuracy comparisons.**

| Completion setting | n | Draft and candidates | Draft only | Candidates only | Question only |
|---|---|---|---|---|---|
| Second model, fixed Qwen drafts | 198 | 156 (78.79%) | 96 (48.48%) | 137 (69.19%) | 54 (27.27%) |
| Qwen, parseable initial submissions | 200 | 192 (96.00%) | 192 (96.00%) | 115 (57.50%) | 61 (30.50%) |
| Qwen, 256-token recovery limit | 198 | 118 (59.60%) | 67 (33.84%) | 92 (46.46%) | 18 (9.09%) |
| Qwen, English recovery prompt | 97 | 57 (58.76%) | 35 (36.08%) | 50 (51.55%) | 7 (7.22%) |

Cells report correct answers and percentages. “Non-triggering” denotes parseable initial submissions, irrespective of initial correctness. The English row uses a 97-problem subset; other rows use Chinese instructions. The second model receives the original Qwen drafts and candidates.

**A second completion model.** On the same 198 problems and fixed materials, the second model gains 9.60 percentage points from drafts with candidates fixed and 30.30 points from candidates with drafts fixed. The paired transitions are 32 versus 13 and 61 versus one, respectively; Holm-adjusted p-values are 0.00661 and $5.46 \times 10^{-17}$. The interaction is −11.62 points (secondary $p = 0.00213$). These results extend the input-effect pattern to a second completion model on the same fixed recovery materials. The API request alias was deepseek-chat; all response identifiers were deepseek-flash.

**Availability of an initial submission.** On 200 non-triggering problems, adding candidates to the supplied draft produces no change in item-level correctness: both conditions answer 192 problems correctly. When the draft is absent, candidates increase the correct count from 61 to 115, a gain of 27 percentage points. Candidate information therefore remains useful in this cohort, with its incremental value depending on whether the previous draft is available. Cohort composition and draft content also differ from the failure cohort, so the between-cohort contrast does not isolate the presence of a final answer.

**Recovery allowance.** Raising the recovery limit from 32 to 256 tokens leaves item-level correctness unchanged across all four conditions. Two output strings do change, reaching 49 and 132 tokens, and parseability improves from 789 to 791 of 792 outputs. Correctness is stable despite these changes in output behavior. Without candidates, a 512-token limit also retains the draft-only and question-only correct counts of 67 and 18, a 24.75-point difference (unadjusted paired $p = 1.36 \times 10^{-10}$). These comparisons retain the single-line submission instruction; they do not test unrestricted continuation of a derivation.

**Prompt language.** With the recovery instructions translated into English, the 97-problem subset retains a positive candidate effect of 22.68 points (Holm-adjusted $p = 2.10 \times 10^{-5}$) and a negative interaction of −21.65 points (secondary p = 0.000402). The draft effect is positive at 7.22 points but is not statistically established (Holm-adjusted p = 0.265; 95% interval [−3.09, 17.53]). Appendix F gives the executed English instructions and same-subset Chinese results. The evidence supports candidate assistance under both instruction languages, while remaining less conclusive for the English draft contrast.

# 5. Discussion

## 5.1 What the comparisons explain

The experiments distinguish two outcomes of an attempt: its submitted answer and the material it leaves for a later call. The development comparison first separates failure to deliver an answer from its mathematical correctness. Under a common submission instruction, the factorial comparison then shows that a later answer depends on which information is carried forward. Because every condition includes a fresh call, the draft gain is not explained by the mere availability of another submission attempt. Because candidates are held fixed in the primary draft contrast, candidate assistance alone does not explain that gain either. The retention comparison adds a further distinction: keeping a similar quantity of text need not preserve a similar amount of downstream usefulness.

Candidate information makes the conditional nature of that value visible. In the failure cohort, it reduces the marginal draft gain while raising overall accuracy. In the cohort with parseable initial submissions, candidates help when the draft is absent but add no correct answers when it is supplied. These outcomes support a common interpretation: an input contributes relative to the information already available, rather than carrying a fixed accuracy benefit of its own.

These findings support an account of answering as a sequence of information-dependent operations. Text produced in one call becomes an input to the next, and its value is expressed through the next call's outcome. Partial calculations, candidate solutions and answer cues may all support that transition. The experiments identify the consequences of retaining or removing the record; they do not establish that the initial model had already internally solved every subsequently recovered problem. This distinction makes the result useful for workflow design without treating the visible draft as a complete explanation of internal computation.

## 5.2 Implications for using language models

For users, the first decision is whether the failure concerns submission. If a derivation ends without a usable answer, a separate request that explicitly asks for the final value is a concrete recovery option. The development comparison shows why the requested deliverable must fit the output allowance; the factorial comparison shows why the existing draft can be worth carrying into that request. A parseable response still needs a correctness check: Algorithm 1 detects missing submissions and does not diagnose a well-formatted but incorrect answer.

The next decision is what information to supply. On the tested failure cohort, retaining the draft adds 26 correct answers on balance when candidates are already present, although it also changes 13 previously correct answers to incorrect ones. This supports evaluating draft reuse as a recovery policy rather than trusting every retained derivation. Candidate availability should also be recorded when judging performance: the tested candidates include the correct value and substantially change recovery accuracy. The result does not justify manufacturing arbitrary options for an open problem and expecting the same improvement.

When a draft must be shortened, the relevant measure is how well the retained material supports the next answer. At similar average lengths, trailing excerpts yield 20 more correct answers than leading excerpts in this cohort. This gives users a concrete retention policy to test on comparable workflows, while the full-versus-trailing result does not establish that shortening is lossless. The methodological lesson is to compare retention choices on the same tasks and final-answer requirements, rather than judge compression by token count alone.

In everyday analysis and enterprise decisions, analogous choices arise when earlier calculations, constraints and alternatives are passed to a later request for a recommendation. The study motivates testing those handoffs: compare the same downstream tasks with and without the prior analysis, then test what survives a shorter record. These are prospective uses of the experimental method. Their success rates must be measured in the target setting, where candidate alternatives may omit a satisfactory answer and correctness may require external judgment.

### 5.3 Scope of the explanation

The strongest evidence concerns paired input changes on a fixed cohort of initially unsuccessful mathematical submissions. A second completion model extends the factorial result on those same drafts; the retention advantage is established in the original Qwen setting. The cohort's source collection had previously been used, and draft classification was automated rather than independently reviewed by two human annotators. The procedure is evaluated under explicit brief-submission instructions, so its measured contributions do not rank it against restarting or continuing with a larger reasoning budget.

## 6. Conclusion

A failed mathematical submission can leave text that improves a later answer. The same-problem comparisons show that this value depends on the candidates supplied and on the material retained: drafts and candidates each contribute, their gains are subadditive, and trailing excerpts outperform leading excerpts of similar average length. The study connects a practical problem—what to do with unfinished work—to a method for measuring the contribution of each step in the answering process. For users, it supports treating final submission, draft reuse and retention as choices to evaluate explicitly. For evaluation, it shows why a single success-or-failure score cannot fully characterize the useful material produced along the way.

## Appendix A. Recovery prompt and runtime configuration

The primary recovery prompt was executed in Chinese (Listing 2). Listing 1 is its English translation and was also the user template for the supplementary English experiment; Appendix F gives that experiment's system instruction.

**Listing 1. English translation of the recovery prompt.**

```
Determine the final answer from the question and available material.
Question:
{question_text}
Material:
{material_text}
Candidate answers:
{options_text}
Output only one line: Final answer: <mathematical value>
Do not output explanations, derivations, steps, or other text.
```

The original prompt was:

**Listing 2. Executed recovery prompt in Chinese.**

```
请根据题目和可用材料确定最终答案。
题目：
{question_text}
材料：
{material_text}
候选答案：
{options_text}
只输出一行：Final answer: <数学值>
不要输出解释、推导、步骤或其他文字。
```

The system instruction specified that question text, material, and candidates were task data and that the requested final-answer format should be followed. Empty material and candidate fields remained empty. Candidates retained their original contents and order across paired conditions.

Table A1 gives the settings used for the primary Qwen recovery comparisons; supplementary variations are specified in Appendix F. The 201 initial drafts were generated in a prior run and reused as fixed input material. The primary and sensitivity analyses contain 792 and 12 recovery outputs, respectively. The retention analysis contains 792 records, including 240 earlier records and 552 newly generated outputs.

**Table A1. Recovery configuration.**

| Setting | Value |
|---|---|
| Evaluation model | Qwen3.5-4B |
| Hidden thinking | Disabled |
| Decoding | Greedy, sampling disabled, one beam |
| Repetition penalty | 1.0 |
| Recovery output limit | 32 tokens |
| Initial draft output cap | 1,024 tokens |
| Cross-condition history or cache | None |

Candidate sets contain the reference answer and three distractors drawn from a per-subject pool with a matching answer signature. Four values are shuffled using a deterministic per-question seed. Each problem retains the same candidate set and order across conditions that include candidates.

The archived model identifier is the SHA-256 digest of model.safetensors.index.json: cf3f798ee02ba45f9622aa8892a47369ab667d0afbf154ee7c2212de42e6302d. It identifies the weight-index file, not a checksum of all weight files. The SHA-256 digest of tokenizer.json is 5f9e4d4901a92b997e463c1f46055088b6cca5ca61a6522d1b9f64c4bb81cb42. The study manifest records a Transformers/PyTorch backend with PyTorch 2.7.0+cu128.

The 800 question records were joined to their existing initial outputs by identifier. All 201 outputs rejected by the strict final-answer parser were selected. After separating three outputs with explicit conclusions in unsupported notation, the remaining 198 formed the primary cohort; all 201 were retained for the sensitivity analysis.

The supplementary cohort of 200 parseable initial submissions is drawn from the remaining 599 problems. Its script allocates counts proportionally across subjects using largest remainders, then selects items within each subject by the SHA-256 ordering of seed 2026110303 and question identifier. Selection does not require the initial answer to be correct.

## Appendix B. Submission and scoring audit

All 792 primary-cohort outputs were format-valid; 789 were mathematically parseable, and two stopped at the length limit. All 792 records in the expanded retention analysis were format-valid and parseable and ended normally. These records are included in the reported denominators.

During the primary run, the parser was extended to support ratio notation and grouped integers. Existing outputs were retained. A later uniform rescore of all 804 primary and sensitivity outputs produced no differences from the stored parseability or correctness labels. This comparison used stored labels, not a rerun of the earliest parser version.

The two primary paired tables are: draft effect, 79 correct in both conditions, 39 correct only with the draft, 13 correct only without it, and 67 incorrect in both; option effect, 62 correct in both, 56 correct only with options, five correct only without options, and 75 incorrect in both. Intervals in the manuscript follow the final audit calculation consistently.

A limited notation check evaluated 248 scoring inputs: the 201 reference answers themselves and 47 additional forms produced by scripted transformations. Of these inputs, 247 were accepted as correct. For 155 problems, the only tested form was the original reference answer. One additional form, `90 - (25)/(2)\pi`, was not accepted against the reference `90 - \frac{25}{2}\pi`. This check documents behavior on those constructed inputs; it does not estimate the error rate on the distribution of model outputs or assess false acceptance of incorrect answers.

Supplementary output validity is reported in Table B1. All assigned outputs remain in each accuracy denominator, including unparseable or format-invalid submissions.

**Table B1. Supplementary submission audit.**

| Experiment | Outputs | Format-valid | Parseable |
|---|---|---|---|
| Second completion model | 792 | 792 | 787 |
| Non-triggering cohort | 800 | 800 | 798 |
| 256-token recovery | 792 | 792 | 791 |
| Candidate reordering | 194 | 194 | 194 |
| Tail edits | 180 | 180 | 180 |
| No candidates, 512 tokens | 396 | 396 | 395 |
| English instructions | 388 | 386 | 387 |

Format validity and parseability are distinct recorded properties; a mathematical value may be extractable from an output that violates the requested format.

## Appendix C. Retention construction and exploratory analyses

The retention study developed in stages. An initial subject-stratified sample of 60 problems used seed 2 026 110 303 and compared full drafts, leading excerpts and no draft. A subsequent trailing condition supplied 60 additional records. The expanded analysis reused these 240 records and generated four conditions for the remaining 138 problems, producing 792 records in total. This is an extension within the primary cohort, not an independent replication on new problems.

Leading excerpts were extracted from the first approximately 600 tokens and shortened to a nearby paragraph or sentence boundary where possible. Trailing excerpts were constructed from the last 600 tokens, trimmed forward to the first paragraph or sentence boundary while retaining at least 450 tokens. Average lengths were 583.8 and 581.7 tokens. The within-problem trailing-minus-leading length difference ranged from −62 to 59 tokens, with an average absolute difference of 13.90 tokens.

**Table C1. Paired retention outcomes on 198 problems.**

| Comparison | Both correct | First only correct | Second only correct | Neither correct |
|---|---|---|---|---|
| Full − trailing | 110 | 8 | 6 | 74 |
| Full − leading | 84 | 34 | 12 | 68 |
| Trailing − leading | 79 | 37 | 17 | 65 |
| Trailing − no draft | 75 | 41 | 17 | 65 |
| Leading − no draft | 77 | 19 | 15 | 87 |
| Full − no draft | 79 | 39 | 13 | 67 |

As a post hoc sensitivity analysis, the 138 problems added in the expansion have 78, 79, 66 and 62 correct answers under full, trailing, leading and empty retention. Trailing minus leading is 9.42 percentage points (23 versus 10 discordant outcomes; unadjusted exact $p = 0.03508$), and trailing minus empty is 12.32 points (28 versus 11; $p = 0.00948$). The corresponding p-values after Holm correction across all six comparisons in this subset are 0.14033 and 0.05685. The directions agree with the combined analysis, without establishing separately significant effects under that correction.

An earlier exploratory intervention compared wording variants and local mutations of leading excerpts. Wording variants preserved recorded mathematical expressions and values, while mutations changed one recorded substring; full semantic effects were not independently adjudicated. Four problems were correct only under the wording variant and one only under mutation (exact $p = 0.375$). Because the two variants were separately constructed and the leading excerpt’s advantage over empty input was not established, this comparison does not identify the content responsible for the full-draft benefit.

## Appendix D. Descriptive candidate-position audit

Table D1 reports accuracy by the position of the reference answer, separately for the two candidate-present conditions. Each row contains different problems, and each problem contributes one outcome to each condition. The initial comparison underlying Table D1 did not change order within problems. The table is descriptive; the subsequent reordering comparison is reported below.

**Table D1. Correct answers by candidate position.**

| Reference-answer position | Draft and candidates | Candidates only |
|---|---:|---:|
| First | 35/48 | 32/48 |
| Second | 31/56 | 19/56 |
| Third | 26/53 | 20/53 |
| Fourth | 26/41 | 21/41 |

Candidate order is fixed within each problem across the draft-present and draft-absent conditions. Their paired contrast consequently does not involve an order change. Absolute accuracies and the estimated benefit of candidates nevertheless pertain to the candidate sets and orders used in this study.

A supplementary intervention reordered the four candidates on 97 problems and repeated the two candidate-present conditions. Accuracy pooled descriptively across those conditions was 55.26%, 64.58%, 36.54% and 53.57% at the new reference-answer positions, compared with 62.50%, 43.48%, 46.15% and 54.17% at the original positions on the same subset. The highest-accuracy position changed, but each position group also changed its problem membership. This pattern alone cannot separate candidate content from positional sensitivity.

The within-problem comparison is more direct. With drafts, reordering changes seven outcomes from incorrect to correct and six from correct to incorrect, increasing the correct count from 56 to 57 of 97. Without drafts, eight outcomes improve and eight worsen, leaving 44 correct. Both exact paired p-values are 1.0. The observed average differences are small, although individual outcomes change; the comparison does not establish that order is irrelevant.

# Appendix E. Exploratory analyses of retained information

## E.1 Joint outcomes of draft and candidate conditions

Table E1 describes all joint outcomes of the three input-present conditions. It shows both successful combinations and cases where supplying both inputs produces an incorrect answer despite a correct answer with one input alone.

**Table E1. Joint correctness on 198 problems.**

| Draft and candidates | Draft only | Candidates only | Problems |
|---|---|---|---:|
| Correct | Correct | Correct | 36 |
| Correct | Correct | Incorrect | 26 |
| Correct | Incorrect | Correct | 43 |
| Correct | Incorrect | Incorrect | 13 |
| Incorrect | Correct | Correct | 2 |
| Incorrect | Correct | Incorrect | 3 |
| Incorrect | Incorrect | Correct | 11 |
| Incorrect | Incorrect | Incorrect | 64 |

Of the 80 problems incorrect with both inputs, 18 are correct in at least one single-input condition. Of the remaining 64, two are correct with the question alone; 62 are incorrect in all four factorial conditions. The categories describe observed responses; failure with both inputs does not imply failure under every configuration.

## E.2 Exploratory matching of reference answers in excerpts

A script searched leading and trailing excerpts for normalized reference-answer strings or matching numerical values. It marked both excerpts on 54 problems, only the trailing excerpt on 29, only the leading excerpt on four, and neither on 111. The procedure uses substring and numerical occurrence checks, rather than semantic identification of an asserted answer. A match can therefore reflect an unrelated number, an exponent or part of a longer number; a missing match can omit an equivalent expression. These counts describe lexical matching, not semantic localization of the information responsible for recovery.

## E.3 Local edits of trailing excerpts

On 60 problems, the original trailing excerpts produced 37 correct answers, paraphrases 35, and local mutations 36. Original versus paraphrased material has two improvements and no reversals when moving back to the original (exact $p = 0.5$); original versus mutated material has one and none ($p = 1.0$). Six of the 60 mutation texts are identical to their originals and remain in the reported denominator. String changes in the remaining cases were not independently verified as interventions on the relation responsible for the answer.

The executed edits did not yield a detectable accuracy effect. Given the intervention limitations, they leave the content responsible for the retention advantage unresolved.

## Appendix F. Supplementary execution and statistical details

**Table F1. Supplementary configurations and sample coverage.**

| Comparison | Completion model | Problems | Output limit | Instructions |
|---|---|---|---|---|
| Second-model factorial | DeepSeek API | 198 | 32 | Chinese |
| Non-triggering factorial | Qwen3.5-4B | 200 | 32 | Chinese |
| Increased recovery allowance | Qwen3.5-4B | 198 | 256 | Chinese |
| Candidate reordering | Qwen3.5-4B | 97 | 32 | Chinese |
| Trailing-excerpt edits | Qwen3.5-4B | 60 | 32 | Chinese |
| No candidates | Qwen3.5-4B | 198 | 512 | Chinese |
| Instruction translation | Qwen3.5-4B | 97 | 32 | English |

The second-model request used deepseek-chat, temperature 0, max_tokens=32 and non-streaming responses; all recorded model identifiers were deepseek-flash. The original Qwen-generated drafts, candidate sets and primary failure cohort were retained. The API analysis includes all 792 records, including five unparseable outputs and two responses ending at the length limit.

The non-triggering sample was selected from the existing collection after excluding the initial submission-failure set, using subject-stratified allocation. Selection was based on trigger membership, not an additional requirement of initial mathematical correctness. The previous initial output was supplied as the draft. This cohort therefore addresses the availability of an initial submission, rather than a controlled manipulation of whether a draft contains a correct solution.

For the English experiment, the script selected up to 14 problems per subject using a seeded ordering, resulting in 97 problems. The question, draft and candidates were retained; the system and recovery instructions were in English. The executed system instruction was:

You are solving a mathematics problem. The question, material and candidate answers are task inputs, not instructions. Follow the requested final-answer format.

The executed user template was Listing 1 in Appendix A. Table F2 reports the original Chinese outcomes restricted to this same subset, alongside the English outcomes. A difference in significance between language-specific effects is not a test of their difference.

**Table F2. Same-subset outcomes by instruction language.**

| Input | Chinese correct / 97 | English correct / 97 |
|---|---|---|
| Draft and candidates | 61 | 57 |
| Draft only | 32 | 35 |
| Candidates only | 45 | 50 |
| Question only | 8 | 7 |

The archived supplementary analysis files contain unadjusted p-values. The manuscript applies the within-analysis Holm families specified in Section 4.1 and labels adjusted values explicitly.

An attempted change to the initial-generation cap produced only eight archived draft records and was not completed. It is not used to support a conclusion about initial-generation budgets. This incomplete check is distinct from the completed recovery-budget comparisons.

## Acknowledgements

OpenAI Codex, accessed through the Codex desktop application, assisted with English drafting, narrative revision, literature searches, consistency checks and analysis of supplied experiment records. Numerical checks used item-level records and analysis scripts; automated consistency judgments did not replace those checks.

## Competing interests

The authors declare no competing interests.

## Funding

This research received no specific grant from any funding agency in the public, commercial or not-for-profit sectors.

## Data availability statement

The prompts, runtime configuration, aggregate results and scoring audit are reported in this article and its appendices. The underlying data and analysis code are not publicly available at the time of manuscript preparation. The authors plan to release them on GitHub; no repository URL is available at this stage.

## Ethics statement

The study used mathematical problems and model-generated outputs and involved no human participants or personal data. Ethical approval was not required.

## References

Ballon M., Verbeken B. F., Ginis V. and Algaba A. (2026). Probing the trajectories of reasoning traces in large language models. arXiv, arXiv:2601.23163. https://doi.org/10.48550/arXiv.2601.23163

Lanham T., Chen A., Radhakrishnan A., et al. (2023). Measuring faithfulness in chain-of-thought reasoning. arXiv, arXiv:2307.13702. https://doi.org/10.48550/arXiv.2307.13702

Liu D., Wang Z., Liu X., Li J., Hu Y., Tang M., Dai B. and Fu J. (2026). When can LLM reasoning reliably reuse intermediate results? arXiv, arXiv:2610.27662. https://doi.org/10.48550/arXiv.2610.27662

Madaan A., Tandon N., Gupta P., et al. (2023). Self-Refine: Iterative refinement with self-feedback. *arXiv*, arXiv:2303.17651. https://doi.org/10.48550/arXiv.2303.17651

Raman N. K., Lundy T. and Leyton-Brown K. (2026). Reasoning models are test exploiters: Rethinking multiple choice. In *Proceedings of the 43rd International Conference on Machine Learning*, Proceedings of Machine Learning Research 306, 103497–518. https://proceedings.mlr.press/v306/raman26a.html

Shinn N., Cassano F., Berman E., Gopinath A., Narasimhan K. and Yao S. (2023). Reflexion: Language agents with verbal reinforcement learning. *arXiv*, arXiv:2303.11366. https://doi.org/10.48550/arXiv.2303.11366

Wang H. and Zhu M. (2026). The detection–extraction gap: Models know the answer before they can say it. *arXiv*, arXiv:2604.06613. https://doi.org/10.48550/arXiv.2604.06613

Wei J., Wang X., Schuurmans D., Bosma M., Ichter B., Xia F., Chi E., Le Q. and Zhou D. (2022). Chain-of-thought prompting elicits reasoning in large language models. *arXiv*, arXiv:2201.11903. https://doi.org/10.48550/arXiv.2201.11903